\def\year{2022}\relax
\documentclass[letterpaper]{article} % DO NOT CHANGE THIS
\usepackage{aaai22}  % DO NOT CHANGE THIS
\usepackage{times}  % DO NOT CHANGE THIS
\usepackage{helvet}  % DO NOT CHANGE THIS
\usepackage{courier}  % DO NOT CHANGE THIS
\usepackage[hyphens]{url}  % DO NOT CHANGE THIS
\usepackage{graphicx} % DO NOT CHANGE THIS
\usepackage{natbib}  % DO NOT CHANGE THIS AND DO NOT ADD ANY OPTIONS TO IT
\usepackage{caption} % DO NOT CHANGE THIS AND DO NOT ADD ANY OPTIONS TO IT
\DeclareCaptionStyle{ruled}{labelfont=normalfont,labelsep=colon,strut=off} % DO NOT CHANGE THIS
\usepackage{subfigure}
\usepackage{stfloats}

\usepackage{hyperref}
\hypersetup{
	colorlinks=true,
	linkcolor=blue,
	filecolor=blue,      
	urlcolor=blue,
	citecolor=black,
}
\usepackage{algorithm}
\usepackage{algorithmic}
\usepackage{amsmath}
\usepackage{amssymb}
\usepackage{multirow}
\usepackage{newfloat}
\usepackage{listings}
\floatstyle{ruled}
\newfloat{listing}{tb}{lst}{}
\floatname{listing}{Listing}
\title{MISSFormer: An Effective Medical Image Segmentation Transformer}
\author{
	Xiaohong Huang,\textsuperscript{\rm 1}
	Zhifang Deng,\textsuperscript{\rm 1}
	Dandan Li,\textsuperscript{\rm 1}\thanks{Corresponding author}
	Xueguang Yuan\textsuperscript{\rm 1}
}
\affiliations{
	\textsuperscript{\rm 1} Beijing University of Posts and Telecommunications\\
	huangxh@bupt.edu.cn, dengzfong@bupt.edu.cn, dandl@bupt.edu.cn, yuanxg@bupt.edu.cn
}

\usepackage{bibentry}
\begin{document}

\maketitle

\begin{abstract}
The CNN-based methods have achieved impressive results in medical image segmentation, but they failed to capture the long-range dependencies due to the inherent locality of the convolution operation. Transformer-based methods are recently popular in vision tasks because of their capacity for long-range dependencies and promising performance. However, it lacks in modeling local context. In this paper, taking medical image segmentation as an example, we present MISSFormer, an effective and powerful Medical Image Segmentation tranSFormer. MISSFormer is a hierarchical encoder-decoder network with two appealing designs: 1) A feed-forward network is redesigned with the proposed Enhanced Transformer Block, which enhances the long-range dependencies and supplements the local context, making the feature more discriminative. 2) We proposed Enhanced Transformer Context Bridge, different from previous methods of modeling only global information, the proposed context bridge with the enhanced transformer block extracts the long-range dependencies and local context of multi-scale features generated by our hierarchical transformer encoder. Driven by these two designs, the MISSFormer shows a solid capacity to capture more discriminative dependencies and context in medical image segmentation. The experiments on multi-organ and cardiac segmentation tasks demonstrate the superiority, effectiveness and robustness of our MISSFormer, the experimental results of MISSFormer trained from scratch even outperform state-of-the-art methods pre-trained on ImageNet. The core designs can be generalized to other visual segmentation tasks. The code has been released on Github:
\href{https://github.com/ZhifangDeng/MISSFormer}{https://github.com/ZhifangDeng/MISSFormer}
\end{abstract}

\section{Introduction}
\noindent With the improvement of medical treatment and the people's health awareness, the requirements of accurate medical image analysis (such as preoperative evaluation, auxiliary diagnosis) have become more critical. The medical image segmentation, as a crucial step of them, the precise and robust segmentation results will provide a sound basis for subsequent analysis and treatment.

Since the fully convolutional networks (FCNs)\cite{Long2015} opened a door for semantic segmentation, one of its variants, the U-shaped networks\cite{2015U,iek20163D} got a promising performance in medical image segmentation by the improvement of skip connection, which provided more detailed information. According to this elegant architecture, the variants of U-Net\cite{Isensee2021,Zhou2018,huang2020unet} have been achieved excellent performance and impressive results. Although their superb performance and prevalence, the CNN-based methods suffer from a limitation in modeling the long-range dependencies because of the locality of convolution operation\cite{Cao2021,Xie2021}, and they failed to achieve the goal of precise medical image analysis. To overcome the limitation, some works proposed dilated convolution\cite{Gu2019,Feng2020} and pyramid pooling\cite{zhao2017pyramid} to enlarge the receptive field as much as possible. And some recent works\cite{Xie2021,mou2019cs,Chen2021,Sinha2020} tried to employ few self-attention layers or transformer layers\cite{vaswani2017attention} in high-level semantic feature maps due to the quadratic relationship between self-attention computational complexity and feature map size, which makes these methods insufficient to capture the abundant long-range dependencies.

Recently, the success of transformers that capture long-range dependencies makes it possible to solve the above problems. Especially, the researches on visual transformer\cite{Liu2021,Dosovitskiy2020,Wang2021,Graham2021,Chu2021,Xie2021a,zheng2021rethinking} are in full swing and have got a promising performance in vision tasks, encouraged by the great success of transformer in natural language processing (NLP). Corresponding to the transformer in NLP, vision transformer\cite{Dosovitskiy2020} fed the image into a standard transformer with positional embeddings by dividing an image into non-overlapping patches and achieved comparable performance with CNN-based methods. Pyramid vision transformer (PVT)\cite{Wang2021} and Swin transformer\cite{Liu2021} proposed hierarchical transformer to explore the vision transformer with spatial reduction attention (SRA) and window-based attention, respectively, which are responsible for reducing computational complexity. Besides, the attempts of SETR\cite{zheng2021rethinking} in semantic segmentation proved the potential of transformer in visual tasks once again. 

However, some recent works\cite{Islam2020,Chu2021a,Li2021} showed the limitation of self-attention on local context, inspired by this, Uformer\cite{Wang2021a}, SegFormer\cite{Xie2021a} and PVTv2\cite{Wang2021b} tried to embed convolutional layer between fully-connected layers of feed-forward network in transformer block to overcome this problem. Despite it captured local context to some extent, but there are some limitations in medical image segmentation: 1) the convolutional layer is embedded between fully-connected layers of the feed-forward network directly, which limits the discrimination of features for our task, although some local context is supplemented, it will be confirmed in Section 4.2; 2) it did not consider the integration of multi-scale information generated by the hierarchical encoder. Both limitations lead to the inferior learning of networks. 

In this paper, MISSFormer, an effective and powerful Medical Image Segmentation tranSFormer, is proposed to leverage the powerful long-ranged dependencies capability of self-attention to produce accurate medical image segmentation. MISSFormer is based on the U-shaped architecture, whose redesigned transformer block, named Enhanced Transformer Block, enhances the feature representations. The MISSFormer consists of encoder, bridge, decoder and skip-connection. These components are all based on the enhanced transformer block. The encoder extracts hierarchical features through the overlapped image patches. Local and global dependencies between different scale features are modeled via the bridge. The decoder is responsible for pixel-wise segmentation prediction with skip connection. The main contributions of this paper can be summarized as follows:
\begin{itemize}
	\item We propose MISSFormer, a position-free and hierarchical U-shaped transformer for medical image segmentation.
	\item We redesign a powerful feed-forward network, Enhanced Mix-FFN, with better feature discrimination, long-range dependencies and local context. Based on this, we expand it and get an Enhanced Transformer Block to make a strong feature representation.
	\item We propose an Enhanced Transformer Context Bridge based on the Enhanced Transformer Block to capture the local and global correlations of hierarchical multi-scale features.
	\item The superior experimental results on medical image segmentation datasets demonstrate the effectiveness, superiority and robustness of the proposed MISSFormer.
\end{itemize}

\section{Related Work}
\textbf{Medical image segmentation.} Medical image segmentation is a pixel-level task of separating the pixels of lesions or organs in a given medical image. U-shaped network\cite{2015U} played a cornerstone role in medical image segmentation tasks because of its superior performance and elegant structure. Due to the rapid development of computer vision tasks\cite{he2016deep,Chen2017}, the medical image segmentation drew lessons from its key insight. For example, resnet architecture became a general encoder backbone for medical image segmentation network, the dilated convolution and pyramid pooling were utilized to enlarge the receptive field for lesion and organ segmentation\cite{Gu2019,Feng2020}. Besides, various attention mechanisms were effective to promote segmentation performance, reverse attention\cite{chen2018reverse} was applied to accurate polyp segmentation\cite{fan2020pranet}, squeeze-and-excitation attention\cite{hu2018squeeze} was integrated into module to refine the channel information to segment vessel in retina images\cite{zhang2019net}, and some works\cite{mou2019cs,Sinha2020} employed self-attention mechanism to supplement the long-range dependencies for segmentation tasks.

\textbf{Vision transformers.} ViT\cite{Dosovitskiy2020} introduced transformer\cite{vaswani2017attention} into visual tasks for the first time and achieved impressive performance because of the capacity for global dependencies of the transformer. Vision tasks developed a new stage inspired by ViT. For example, DeiT\cite{touvron2021training} explored the efficient training strategies for ViT, PVT\cite{Wang2021} proposed a pyramid transformer with SRA to reduce the computational complexity, and Swin transformer\cite{Liu2021} was an efficient and effective hierarchical vision transformer, whose window-based mechanism enhances the locality of features, which was also the improvement of some excellent transformer works\cite{Islam2020,Chu2021a,Li2021}. For other specific tasks, SETR\cite{Zheng2021} was a semantic segmentation network based on the transformer and made ViT as a backbone, SegFormer\cite{Xie2021a} introduced a simple and efficient design for semantic segmentation powered by transformer, DETR\cite{Carion2020} proposed an end-to-end object detection framework with transformer, Uformer\cite{Wang2021a} built a general U-shaped transformer for image restoration.

\textbf{Transformers for medical image segmentation.} Researchers borrowed the transformer to medical image segmentation inspired by the rapid development of vision transformers. Transunet\cite{Chen2021} employed some transformer layers into the low-resolution encoder feature maps to capture the long-range dependencies, UNETR\cite{Hatamizadeh2021} applied transformer to make a powerful encoder for 3d medical image segmentation with CNN decoder, CoTr\cite{Xie2021} and TransBTS\cite{Wang2021c} bridged the CNN-based encoder and decoder with the transformer to improve the segmentation performance in low-resolution stage. Besides these methods which are the combination of CNN and transformer, \cite{Cao2021} proposed Swin-Unet, based on Swin transformer\cite{Liu2021}, to demonstrate the application potential of pure transformer in medical image segmentation. However, Swin-Unet, whose encoder backbone is Swin transformer pre-trained on ImageNet, requires pre-training on large-scale datasets. Different from it, the proposed MISSFormer is trained on the medical image datasets from scratch and achieves better performance because of the discriminative feature representations by Enhanced Transformer Block. 
\begin{figure*}[t]
	\centering
	\includegraphics[width=2\columnwidth]{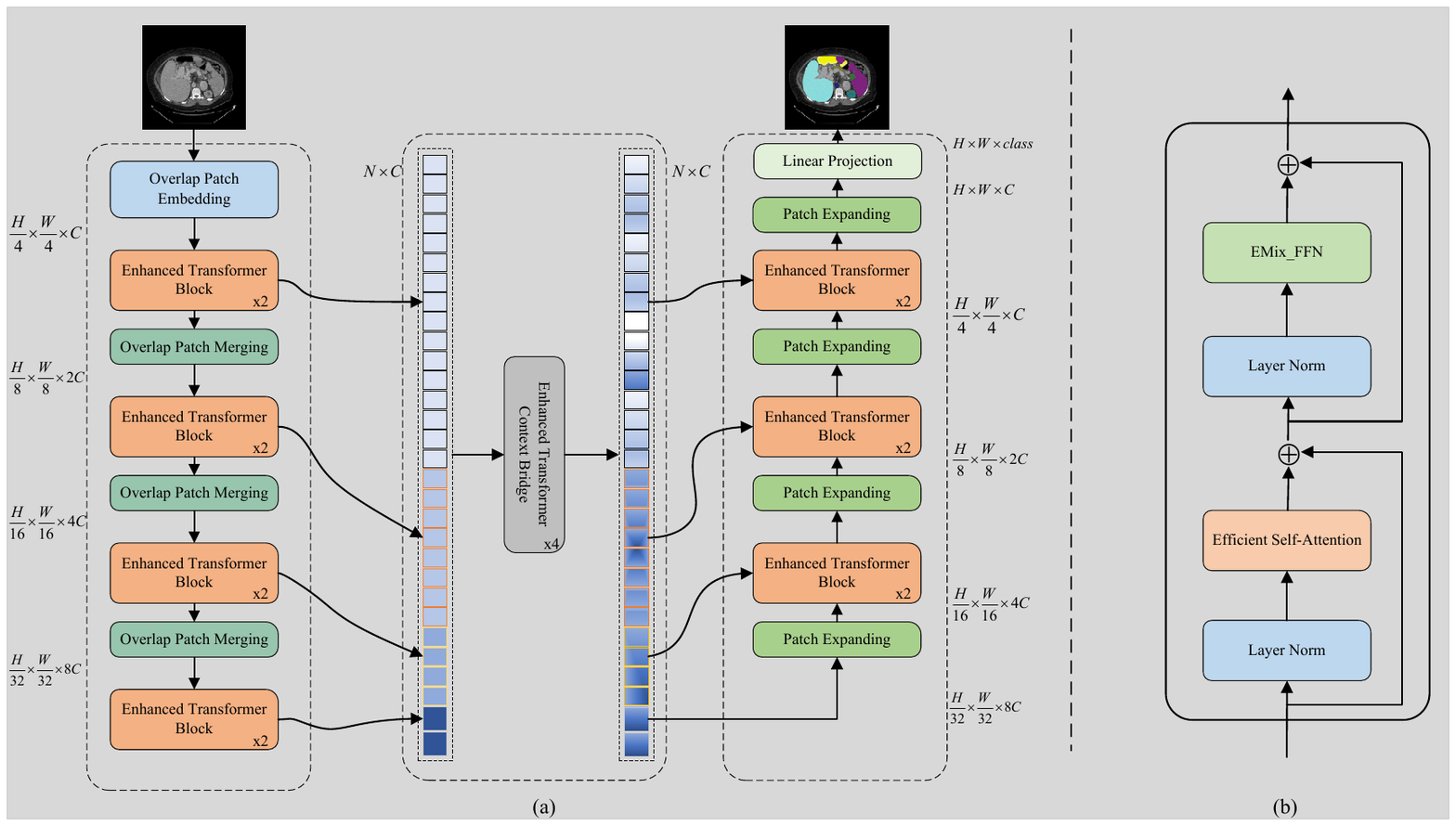} % Reduce the figure size so that it is slightly narrower than the column. Don't use precise values for figure width.This setup will avoid overfull boxes.
	\caption{The overall structure of the proposed MISSFormer. (a) The proposed MISSFormer framework. (b) The structure of Enhanced Transformer Block.}
	\label{fig1}
\end{figure*}
\section{Method}
This section describes the overall pipeline and the specific structure of MISSFormer first, and then we show the details of the improved transformer block, Enhanced Transformer Block, which is the basic unit of MISSFormer. After that, we introduce the proposed Enhanced Transformer Context Bridge, which models the local and global correlations of hierarchical multi-scale information.
\subsection{Overall Pipeline}
The proposed MISSFormer is shown in Fig.1(a), a hierarchical encoder-decoder architecture with an enhanced transformer context bridge module appended between encoder and decoder. Specifically, given an input image, MISSFormer first divides it into overlapping patches of size 4*4 to preserve its local continuity with convolutional layers. Then, the overlapping patches are fed into the encoder to produce the multi-scale features. Here, the encoder is hierarchical, and each stage includes enhanced transformer blocks and patch merging layer. The enhanced transformer block learns the long-range dependencies and local context with limited computational complexity, patch merging layer is applied to generate the downsampling features.

After that, MISSFormer makes the generated multi-scale features pass through the Enhanced Transformer Context Bridge to capture the local and global correlations of different scale features. In practice, different level features are flattened in spatial dimension and reshaped to make consistent in channel dimension, then concatenate them in flattened spatial dimension and feed into the enhanced transformer context bridge with \textit{d-}depth. After that, we split and restore them to their original shape and obtain the discriminative hierarchical multi-scale features. 

For the segmentation prediction, MISSFormer takes the discriminative features and skip connections as inputs of decoders. Each decoder stage includes Enhanced Transformer Blocks and patch expanding layer\cite{Cao2021}. Contrary to the patch merging layer, the patch expanding layer upsample the adjacent feature maps to twice the original resolution except that the last one is four times. Last, the pixel-wise segmentation prediction is output by a linear projection.

\begin{figure*}[t]
	\centering
	\includegraphics[width=2\columnwidth]{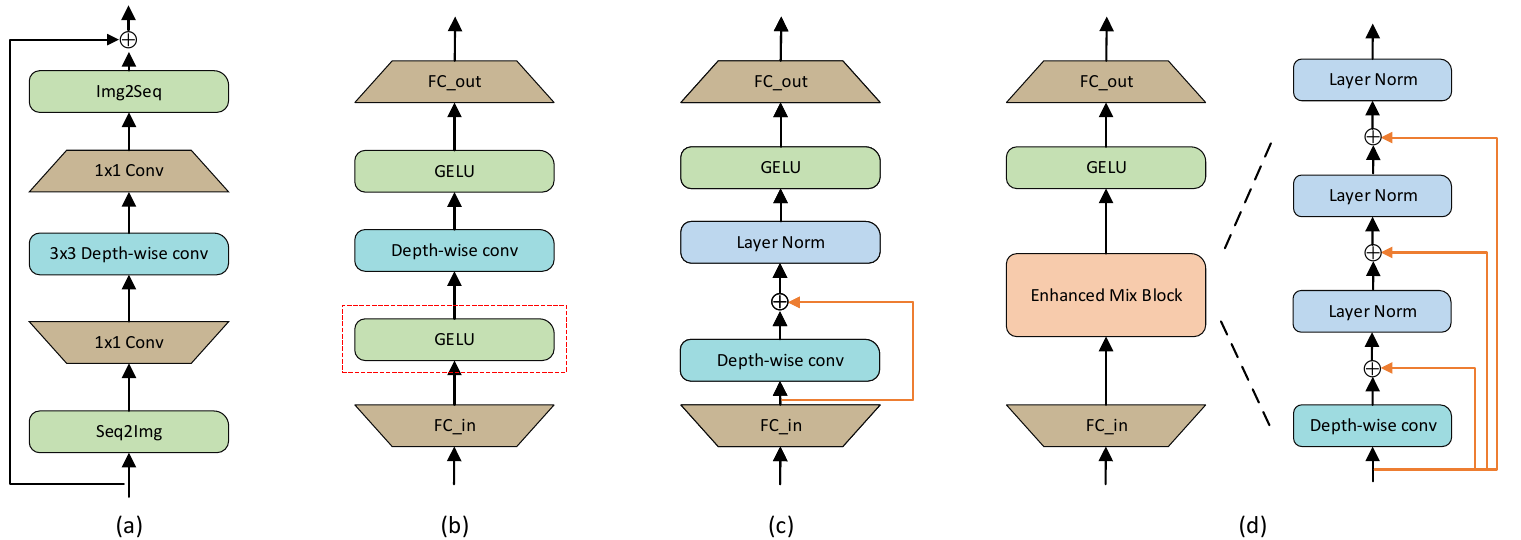} % Reduce the figure size so that it is slightly narrower than the column. Don't use precise values for figure width.This setup will avoid overfull boxes.
	\caption{The various exploration of locality in feed-forward neural network, from left to right: (a)Residual Block in LocalViT, (b)LeFF in Uformer, Mix-FFN in SegFormer and PVTv2, (c) proposed Simple Enhanced Mix-FFN, (d) proposed Enhanced Mix-FFN}
	\label{fig2}
\end{figure*}
\subsection{Enhanced Transformer Block}
Long-range dependencies and local context are effective for accurate medical image segmentation. Transformer and convolution are good choices for long-range dependencies and locality at present, respectively. At the same time, the computational complexity of the original transformer block is quadratic with the feature map resolution, making it unsuitable for high-resolution feature maps. Second, the transformer lacks the ability to extract the local context\cite{Islam2020,Chu2021a,Li2021}, although Uformer, SegFormer and PVTv2 tried to overcome the limitation by embedding a convolutional layer in feed-forward network directly, we argue that this approach limits the discrimination of features, even some improved performance is achieved by them. 

To solve the above problems, we proposed Enhanced Transformer Block. As is shown in Fig.1(b), the Enhanced Transformer Block is composed of LayerNorm, Efficient Self-Attention and Enhanced Mix-FFN.

\textbf{Efficient Self-Attention.} Efficient self-attention is a spatial reduction self-attention\cite{Wang2021}, which can be applied to high-resolution feature map. Given a feature map $\textsl{$F$}$$\in \mathbf{R}^{H\times W\times C}$, and $\textsl{$H,W,C$}$ is the height, width and channel depth respectively. For the original standard multi-head self-attention, it makes $\textsl{$Q,K,V$}$ have same shape $\textsl{$N \times C$}$, where $\textsl{$N = H \times W$}$, which can be formulated as:
\begin{equation}
	Attention(Q, K, V) = {SoftMax(\frac{QK^T}{\sqrt{d_{head}}})}V, 
\end{equation}
and its computational complexity is $\mathcal{O}(N^2)$. While for the efficient self-attention, it applied a spatial reduction ratio $R$ to  reduce the spatial resolution as follows:
\begin{equation}
	{new\_K} = Reshape(\frac{N}{R},C\cdot R)W(C\cdot R,C),
\end{equation}
it first reshapes $K$ and $V$ to $\frac{N}{R}\times (C\cdot R)$, and then a linear projection $W$ is used to make channel depth restore to $C$. After that, the computational complexity of self-attention reduces to $\mathcal{O}(\frac{N^2}{R})$, and can be applied to high-resolution feature maps. The spatial reduction operation is convolution or pooling in common.

\textbf{Enhanced Mix-FFN.} Different from previous methods in Fig.2(a) and (b), we redesigned the structure of Mix-FFN to align features and make discriminative representations. As shown in Fig.2(c), First we add a skip connection before the depth-wise convolution, and then, we applied layer norm after the skip connection, which can be denoted as:
\begin{equation}
\begin{aligned}
	&y_1 = \textbf{LN}(Conv_{3\times3}(FC(x_{in}))+ \textbf{FC}(x_{in})),\\
	&x_{out} = FC(GELU(y_1))+x_{in},
\end{aligned}
\end{equation}
where, $x_{in}$ is the output of efficient self-attention, $Conv_{3\times3}$ is convolution with kernel $3\times3$, we applied depth-wise convolution for efficiency in this paper. We will show that these improvements are essential for Mix-FFN in Section 4.2.

Inspired by\cite{liu2020rethinking}, we extend our design to a general form with the help of layer norm, which facilitates the optimization of skip connection\cite{vaswani2017attention}. As shown in Fig.2(d), we make an Enhanced Mix block embedded in the original feed-forward network. We introduced recursive skip connection in Enhanced Mix block, given an input feature map $x_{in}$, a depth-wise convolution layer is applied to capture the local context, and then a recursive skip connection followed, and it can be defined as:
\begin{equation}
	\begin{aligned}
	&y_i = LN(x_{in}+y_{i-1}),\\
	&x_{out} = FC(GELU(y_i))+x_{in},
	\end{aligned}
\end{equation}
where $ y_1 = LN(x_{in}+F(x_{in},W)) $.
After that, the model makes more expressive power due to the construction of different feature distribution and consistency by each recursive step.

\subsection{Enhanced Transformer Context Bridge}

\begin{figure}[t]
	\centering
	\includegraphics[width=\columnwidth]{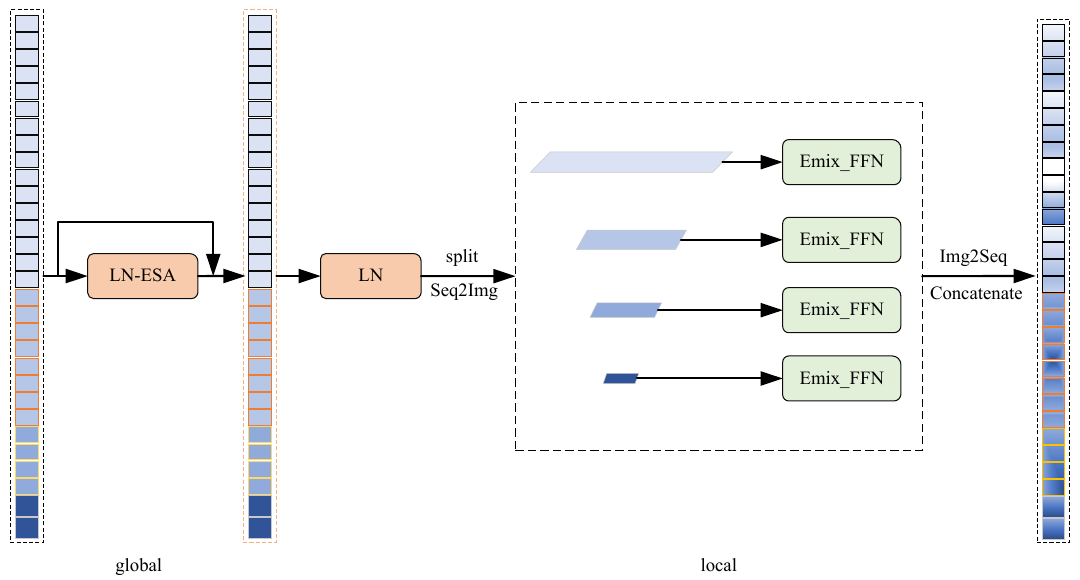} % Reduce the figure size so that it is slightly narrower than the column. Don't use precise values for figure width.This setup will avoid overfull boxes.
	\caption{The Enhanced Transformer Context Bridge}
	\label{fig3}
\end{figure}

Multi-scale information fusion has been proved to be crucial for accurate semantic segmentation in the CNN-based method\cite{Sinha2020,Chen2017}. In this part, we explore the multi-scale feature fusion for the Transformer-based method with the aid of the hierarchical structure of MISSFormer. As is shown in Fig.3, the multiple stage feature maps are obtained after feeding the patches into the encoder, whose settings of patch merging and channel depth in each stage keep the same with SegFormer. Given multi-level features ${F_1, F_2, F_3, F_4}$, which are generated by hierarchical encoder, we flatten them in spatial dimension and reshape them to keep the same channel depth with each other, then we concatenate them in the flattened spatial dimension, after that, the concatenated token is fed into enhanced transformer block to construct the long-range dependencies and local context correlation. The process can be summarized as formula (5).

\begin{equation}
	\begin{aligned}
		&token\_F_{i} = Reshape(F_i,[B,-1,C])\\
		&merge\_token = Concatenate(token\_F_i,dim=1)\\
		&Atten\_token = Efficient\_Atten(LN(merge\_token))\\
		&res\_token = LN(Atten\_token+merge\_token)	\\
		&split\_token = Split(res\_token, dim=1)\\
		&FFN_i = EnhancedMix-FFN(split\_token)\\
		&output = Concatenate(FFN_i,dim=1)+res\_token
	\end{aligned}	
\end{equation}

After the feature passes through $d$ enhanced transformer block, we split tokens, restore them to the original shape of features in each stage, and feed them into a transformer-based decoder with a corresponding skip connection to predict the pixel-wise segmentation map. The depth of Context Bridge is set to 4 in this paper.
\section{Experiments}
In this section, we first conduct the experiment of ablation studies to validate the effectiveness of each component in MISSFormer, and then the comparison results with previous state-of-the-art methods are reported to demonstrate the superiority of the proposed MISSFormer.
\subsection{Experiments Settings}
\textbf{Datasets.} We perform experiments on two different formats of datasets: Synapse multi-organ segmentation dataset (Synapse) and Automated cardiac diagnosis challenge dataset (ACDC). The Synapse dataset includes 30 abdominal CT scans with 3779 axial abdominal clinical CT images, and the dataset is divided into 18 scans for training and 12 for testing randomly, follow the \cite{Cao2021,Chen2021}. We evaluate our method with the average Dice-Sørensen Coefficient (DSC) and average Hausdorff Distance (HD) on 8 abdominal organs (aorta, gallbladder, spleen, left kidney, right kidney, liver, pancreas, spleen, stomach). The ACDC dataset includes 100 MRI scans collected from different patients, and each scan labeled three organs, left ventricle (LV), right ventricle (RV) and myocardium (MYO). Consistent with the previous method\cite{Cao2021,Chen2021}, 70 cases are used for training, 10 for validation and 20 for testing, and the average DSC is applied to evaluate the method.

\textbf{Implementation details.} The MISSFormer is implemented based on PyTorch and trained on Nvidia GeForce RTX 3090 GPU with 24 GB memory. Different from previous work\cite{Cao2021,Chen2021}, whose model is initialized by the pre-trained model on ImageNet, the MISSFormer is initialized randomly and trained from scratch, so the moderate data augmentation is conducted for all datasets. We set the input image size as 224{$\times$}224, the initial learning rate is 0.05 and poly learning rate policy is used, the max training epoch is 400 with a batch size of 24. SGD optimizer with momentum 0.9 and weight decay 1e-4 is adopted for MISSFormer. 
\begin{table}[t]
	\centering
	\begin{tabular}{l|cc}	
		\hline
		Architecture &DSC$\uparrow$ &HD$\downarrow$ \\
		\hline
		\hline
		SegFormer\_B1 (Xie et al. 2021) &75.24 &25.07 \\
		SegFormer\_B5  &75.74&\textbf{21.62}  \\
		U-mlpFormer  &75.88 &27.22 \\
		U-SegFormer& \textbf{76.10}&26.97\\
		\hline
	\end{tabular}
	\caption{Accuracy on Synapse dataset of SegFormer and U-SegFormer}
	\label{table1}
\end{table}
\begin{table}[t]
	\centering
	\begin{tabular}{l|cccc}
		\hline
		Architecture & skip&  LN&DSC$\uparrow$ &HD$\downarrow$\\
		\hline
		\hline
		U-SegFormer& --& --& 76.10 &26.97 \\
		U-SegFormer w/skip& cat& --& 78.14&28.77\\
		U-SegFormer w/skip& add& --& 78.74&20.20\\
		Simple\_MISSFormer& \checkmark& \checkmark& \textbf{79.73}&\textbf{20.14}\\
		\hline			
	\end{tabular}
	%}
	\caption{Effectiveness of Simple Enhanced Mix-FFN, cat denotes concatenation operation for skip connection, add denotes summation. }
	\label{table2}
\end{table}

\subsection{Ablation Studies}
We conduct ablation studies on the Synapse dataset to varify the effectiveness of the essential component in our approach. We set the SegFormer\_B1 as the baseline method, and the number of transformer blocks in every stage of encoder and decoder is set to 2 to keep the same with other methods for a fair comparison. All experiments are performed with the same super parameter settings and trained from scratch. 

\textbf{Architecture selection.} We replace the SegFormer\_B1 MLP decoder with its transformer block and patch expanding to make it U-shaped SegFormer, called ``U-SegFormer", and the results are shown in Table 1. As we can see, the U-SegFormer achieved better performance than SegFormer because the U-shaped model can fuse more corresponding details information with skip connection in each stage, although the SegFormer integrates multi-level information. In order to verify the priority of the U-shaped structure, the naive U-shaped transformer with original mlp FFN and SegFormer\_B5 are also used as comparison results. SegFormer\_B5 has not achieved breakthrough results because of the limitation of a huge number of parameters and medical dataset size. The results of Table 1 demonstrate the priority of U-shaped network structure based on the transformer for medical image segmentation as before.

\begin{figure*}[htbp]
	\centering
	\subfigure[U-mlpFormer.]{
		\begin{minipage}[t]{0.3\linewidth}
			\centering
			\includegraphics[width=2in]{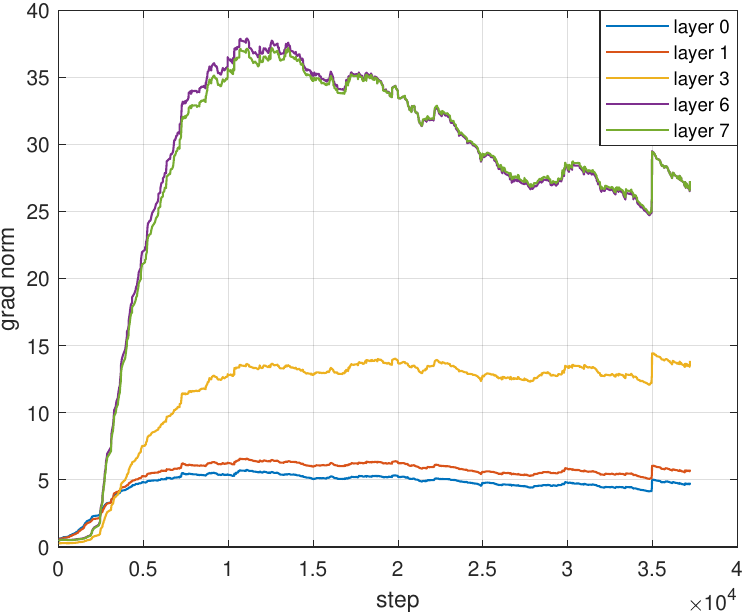}
			%\caption{fig1}
		\end{minipage}%
	}%
	\subfigure[U-SegFormer.]{
		\begin{minipage}[t]{0.3\linewidth}
			\centering
			\includegraphics[width=2in]{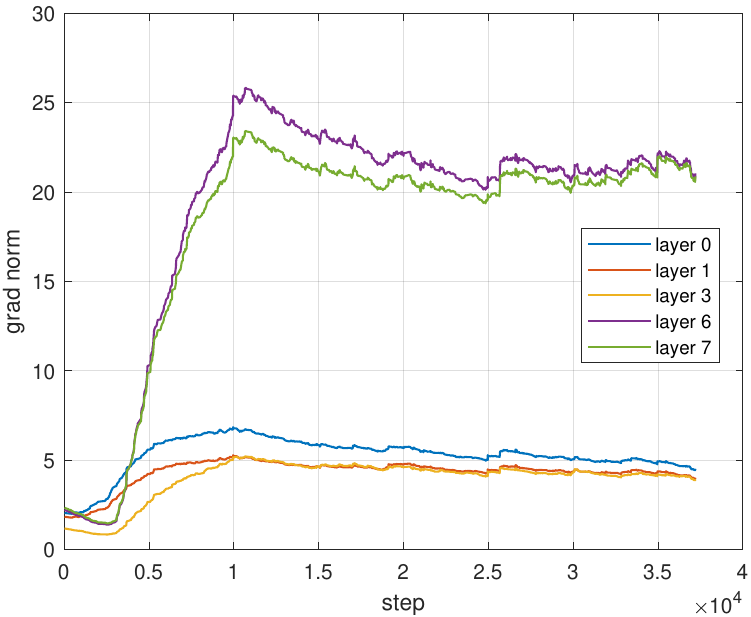}
			%\caption{fig2}
		\end{minipage}%
	}%
	\subfigure[U-SegFormer w/skip.]{
		\begin{minipage}[t]{0.3\linewidth}
			\centering
			\includegraphics[width=2in]{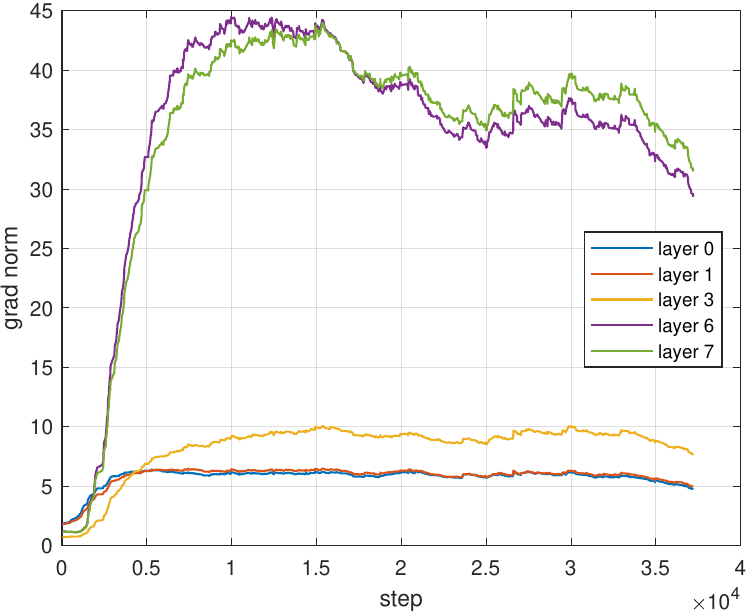}
			%\caption{fig2}
		\end{minipage}
	}%

	\subfigure[U-SegFormer w/LN.]{
		\begin{minipage}[t]{0.3\linewidth}
			\centering
			\includegraphics[width=2in]{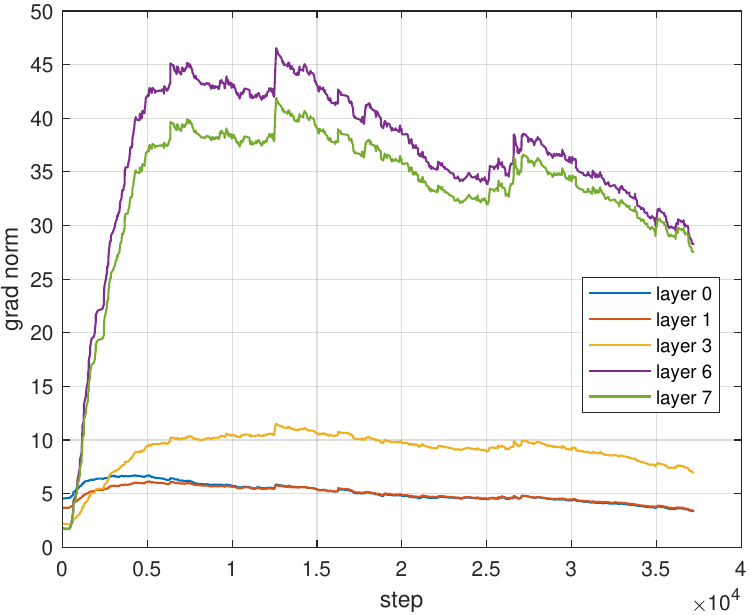}
			%\caption{fig2}
		\end{minipage}
	}%
	\subfigure[Simple\_MISSFormer.]{
		\begin{minipage}[t]{0.3\linewidth}
			\centering
			\includegraphics[width=2in]{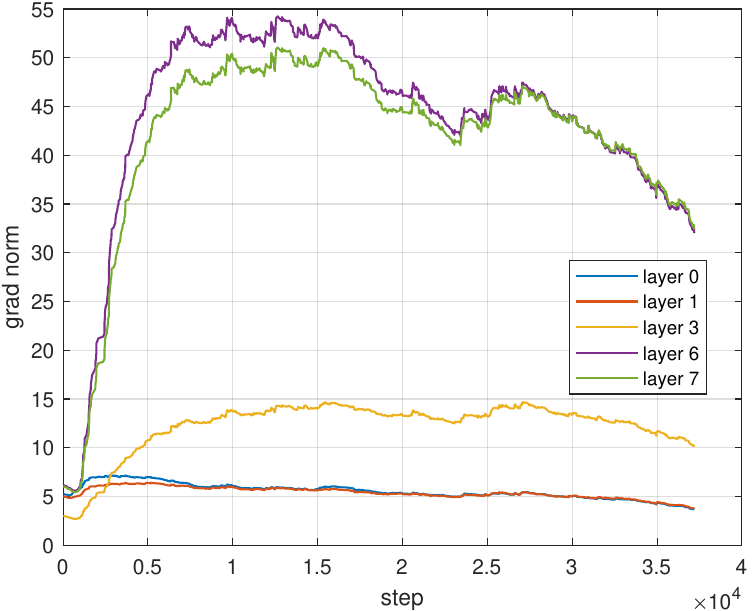}
			%\caption{fig2}
		\end{minipage}
	}%
    \subfigure[MISSFormer.]{
    	\begin{minipage}[t]{0.3\linewidth}
    		\centering
    		\includegraphics[width=2in]{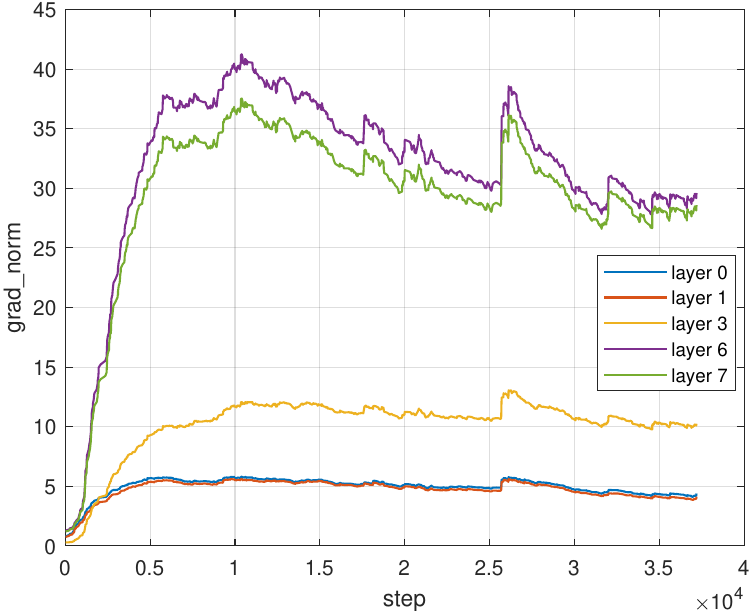}
    		%\caption{fig2}
    	\end{minipage}
    }%
	\centering
	\caption{The average L1 norm of gradients to the second fully connected weight in FFN for layer 0,1,3,6,7}
\end{figure*}

\begin{figure}[]
	\centering
	\subfigure[total\_loss.]{
		\begin{minipage}[t]{0.6\linewidth}
			%\centering
			\includegraphics[width=2in]{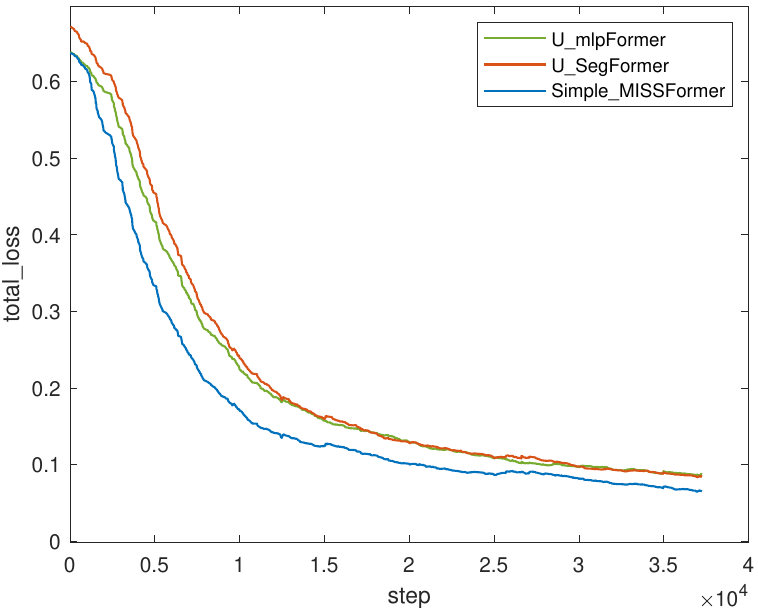}
			%\caption{fig1}
		\end{minipage}%
	}%

	\subfigure[dice.]{
		\begin{minipage}[t]{0.6\linewidth}
			%\centering
			\includegraphics[width=2in]{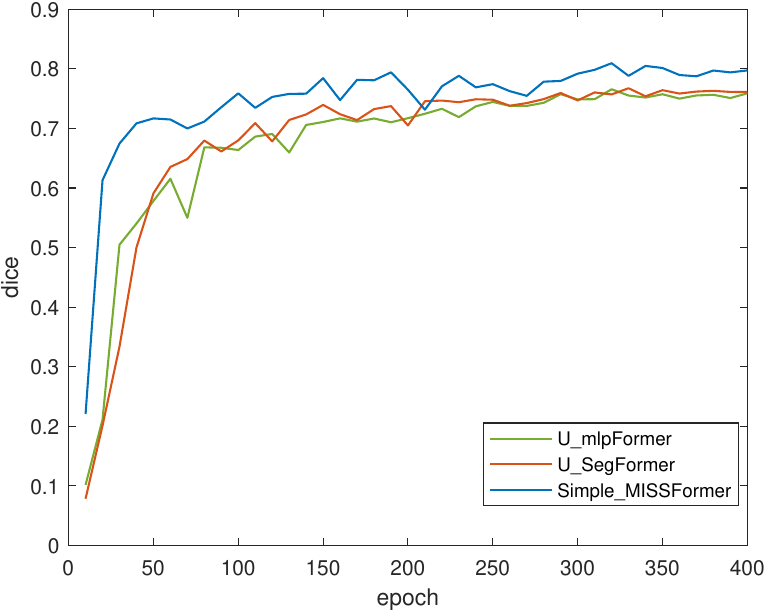}
			%\caption{fig2}
		\end{minipage}%
	}%
	%\centering
	\caption{The convergence and evaluation results of different methods.}
\end{figure}
\textbf{Ablation studies of Simple Enhanced Mix-FFN.} Based on the U-shaped transformer, we further perform the experiments to validate the impact of the proposed Simple Enhanced Mix-FFN components. Table 2 reports the comparison results. We first design different skip connections: concatenation and summation. Table 2 shows that both skip connections boost the model performance considerably, and the summation skip connection improved more than 2.6\%. Then, we explore the gap caused by the direct embedding of convolution. A layer norm is applied to align the feature and distribution. We integrate it after the skip connection, which is called Simple  MISSFormer, and it has 1\% improvements based on U-SegFormer w/skip. Finally, with the help of the redesigned feed-forward network, we improved feature distributions and enhanced feature representations to generate an increasing promotion of 3.63 DSC, compared with the U-SegFormer baseline.

\textbf{Analysis of Simple Enhanced Mix-FFN.} In order to explore the reasons for the effectiveness of the above improvements, we observe the tendency of gradients to the second fully connected weight in FFN for different models with 8 encoder layers. Fig.4 shows the average L1 norm of gradients of different layers of models. We observe that the gradients to the middle layer in U-SegFormer become 1/3 compared with U-mlpFormer, which indicates that direct embedding of 3$\times$3 convolution between fully connected layers makes the update of the middle layer slow and may not get better weights, although it supplements local information and makes slight improvements. In comparison, the two proposed components can solve this problem and make better convergence and evaluation results, as shown in Fig.5.
\setcounter{table}{7}
\begin{table*}[t]
	\centering
	\begin{tabular}{l|c|c|c|c|c|c|c|c|c|c}
		\hline
		Methods & DSC$\uparrow$ & HD $\downarrow$ & Aorta & Gallbladdr&Kidney(L)&Kidney(R)&Liver&Pancreas&Spleen&Stomach \\
		\hline
		\hline
		%\cline{2-4}
		V-Net&68.81&-&75.34&51.87&77.10&80.75&87.84&40.05&80.56&56.98\\
		DARR &69.77&-&74.74&53.77&72.31&73.24&94.08&54.18&89.90&45.96\\
		R50 U-Net&74.68&36.87&87.74&63.66&80.60&78.19&93.74&56.90&85.87&74.16\\
		U-Net&76.85&39.70&89.07&\textbf{69.72}&77.77&68.60&93.43&53.98&86.67&75.58\\
		R50 Att-Unet&75.57&36.97&55.92&63.91&79.20&72.71&93.56&49.37&87.19&74.95\\
		Att-UNet&77.77&36.02&\textbf{89.55}&68.88&77.98&71.11&93.57&58.04&87.30&75.75\\
		R50 ViT&71.29&32.87&73.73&55.13&75.80&72.20&91.51&45.99&81.99&73.95\\
		Transunet&77.48&31.69&87.23&63.13&81.87&77.02&94.08&55.86&85.08&75.62\\
		Swin-Unet&79.13&21.55&85.47&66.53&83.28&79.61&94.29&56.58&90.66&76.60\\
		\hline
		MISSFormer\_S&80.74&19.65&85.31&66.47&83.37&81.65&\textbf{94.52}&63.49&91.51&79.63\\
		MISSFormer&\textbf{81.96}&\textbf{18.20}&86.99&68.65&\textbf{85.21}&\textbf{82.00}&94.41&\textbf{65.67}&\textbf{91.92}&\textbf{80.81}\\
		\hline
	\end{tabular}
	\caption{Comparison with state-of-the-art methods on Synapse dataset. The results of other experiments are original from Swin-Unet\cite{Cao2021}.}
	\label{table8}
\end{table*}
\setcounter{table}{2}
\begin{table}[!h]
	\centering
	\begin{tabular}{l|cc}	
		\hline
		Architecture &DSC$\uparrow$ &HD$\downarrow$ \\
		\hline
		\hline
		U-mlpFormer  &75.88 &27.22 \\
		U-SegFormer& 76.10&26.97\\
		U-LocalViT& 76.92&23.62\\
		Simple\_MISSFormer & \textbf{79.73}& \textbf{20.14}\\
		\hline
	\end{tabular}
	\caption{Comparison of different methods to supplement local information}
	\label{table3}
\end{table}
\textbf{Comparison of different methods to supplement local information.} In order to prove the necessity of supplementing local information and the effectiveness of the proposed method, we compare it with other methods of supplementing local information. Keeping the U-shaped structure unchanged, the experiment is carried out by replacing the FFN in the transformer block with different modules, such as Mix-FFN in SegFormer, residual blocks in LocalViT and the proposed Enhanced Mix-FFN in Simple\_MISSFormer. The result is shown in Table 3.

\begin{table}[!h]
	\centering
	\begin{tabular}{l|ccc}
		\hline
		Achitecture & step &DSC$\uparrow$& HD $\downarrow$ \\
		\hline
		\hline
		%\cline{2-4}
		\multirow{3}{*}{MISSFormer\_S}
		& 1& 79.73 &20.14	\\
		& 2& 79.91 &21.33	\\
		& 3& \textbf{80.74} & \textbf{19.65}	\\
		\hline	
	\end{tabular}
	\caption{Impact of recursive skip connection in Enhanced Mix-FFN, step means recursive step. }
	\label{table4}
\end{table}
\textbf{Impact of further feature consistency in Enhanced Mix-FFN.} Inspired by the above exploration and \cite{liu2020rethinking}, we extend the redesigned FFN of Simple\_MISSFormer to make it more general. We call it MISSFormer\_S due to the absence of multi-scale feature integration. We design experiments to assess the influence of further consistency and distribution caused by different recursive steps, and its results are recorded in Table 4. The results improve with the increase of the recursive step, which further improved the insufficient feature discrimination when convolution is embedded directly in FFN.

\begin{table}[h!]
	\centering
	\begin{tabular}{l|cccc}
		\hline
		Achitecture & step& bridge\_4 & DSC$\uparrow$ & HD $\downarrow$ \\
		\hline
		\hline
		%\cline{2-4}
		\multirow{4}{*}{MISSFormer\_S}
		& 1&--& 79.73 &20.14	\\
		& 2&--& 79.91 &21.33	\\
		& 3&--& 80.74 &19.65	\\	
		\hline
		\multirow{4}{*}{MISSFormer}
		& 1&\checkmark& \textbf{81.96} &\textbf{18.20}	\\
		& 2&\checkmark& 80.91 &19.48	\\
		& 3&\checkmark& 80.72 &23.43	\\	
		\hline
	\end{tabular}
	\caption{Impact of Enhance Transformer Context Bridge on recursive skip connection of MISSFormer. }
	\label{table5}
\end{table}

\begin{table}[h!]
	\centering
	\begin{tabular}{l|cccc}
		\hline
		Achitecture & depth & stage & DSC$\uparrow$ & HD $\downarrow$  \\
		\hline
		\hline
		%\cline{2-4}
		\multirow{6}{*}{MISSFormer    }
		& 2&4/3/2/1& 80.19 &18.88	\\
		& 4&4/3/2/1& \textbf{81.96} &\textbf{18.20}	\\
		& 6&4/3/2/1&81.03 &21.36\\	\cline{2-5} 
		%\hline
		%\multirow{3}{*}{MISSFormer}
		& 4&4/3/2& 80.65&18.39\\
		& 4&4/3& 79.86 &20.33	\\
		& 4&4& 79.56 &20.95 \\	
		\hline
	\end{tabular}
	%}
	\caption{Exploration of the bridge depth and multi-scale information in MISSFormer. }
	\label{table6}
\end{table}

\textbf{Influence of Enhanced Transformer Context Bridge.} We conducted experiments to explore the role of multi-scale information in transformer-based methods on account of the hierarchical features generated by the MISSFormer encoder. As Table 5 shows, we list the results of MISSFormer\_S for intuitionistic comparison, and the performance of the model has been improved to varying degrees except step equals 3 after embedding the Enhanced Transformer Context Bridge into MISSFormer\_S, we call it as MISSFormer. We observe that the model achieved the best performance to have a 2.26\% DSC improvement when the step is 1 and the growth rate gradually decreases with the increase of recursive step, even negative. We guess there is a balance between the recursive step and Enhanced Transformer Context Bridge or between the number of layer norm and model capacity, which will be discussed in our future work. Besides, we also investigated how the bridge depth and multi-scale information integration affect model performance, and the results are saved in Table 6. For the exploration of bridge depth, 4 is a suitable depth in MISSFormer because of the limited medical data. For transformer-based hierarchical features, the more scale features are fed into the enhanced transformer context bridge, the more comprehensive the model can be learned for long-range dependencies and local context.

\begin{figure*}[t]
	\centering
	\includegraphics[width=2\columnwidth]{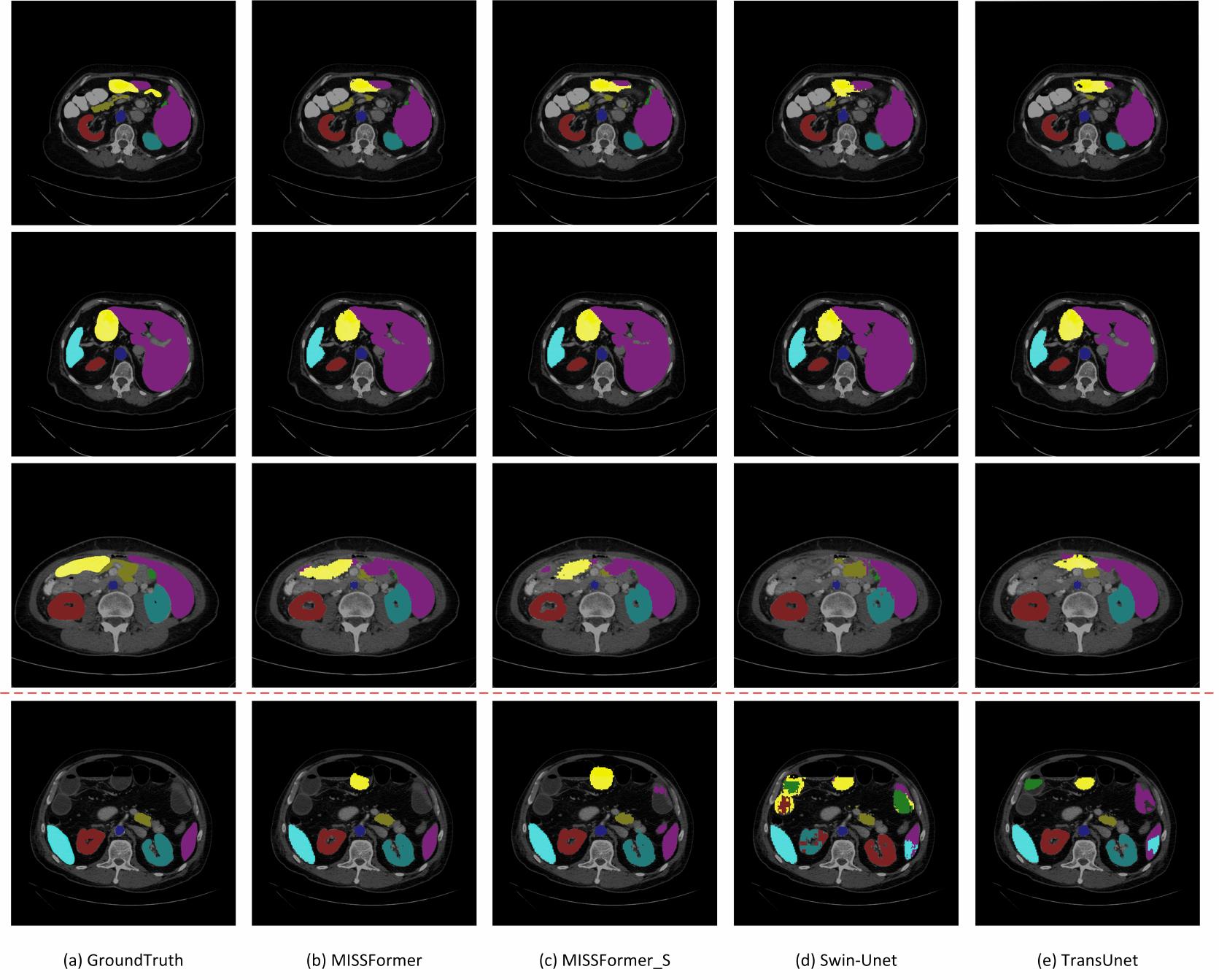} % Reduce the figure size so that it is slightly narrower than the column. Don't use precise values for figure width.This setup will avoid overfull boxes.
	\caption{The visual comparison with previous state-of-the-art methods on Synapse dataset. Above the red line is good cases, and below it is a failed case, Our MISSFormer shows a better performance than other methods}.
	\label{fig3}
\end{figure*}

\begin{table}[!h]
	\centering
	\begin{tabular}{l|c|c|c}
		\hline
		Achitecture & Context Bridge &DSC$\uparrow$ & HD $\downarrow$ \\
		\hline
		\hline
		%\cline{2-4}
		\multirow{4}{*}{MISSFormer}
		& no& 79.73 &20.14	\\
		& mlp& 79.54 &\textbf{17.26}	\\
		& Mix\_FNN& 80.18 &20.17	\\
		& Enhanced Mix\_FFN& \textbf{81.96}&18.20	\\
		\hline	
	\end{tabular}
	\caption{Comparison of different modules in Transformer Context Bridge. }
	\label{table7}
\end{table}

\textbf{The necessity of global-local information in Transformer Context Bridge.} To further explore the effectiveness of the proposed module and the impact of each feature component in multi-scale information aggregation, we take MISSFormer with the depth of 4 as a basis and replace the FFN module in Transformer Context Bridge as mlp\_FFN, Mix\_FFN and Enhanced Mix\_FFN, respectively. The results are recorded in Table 7. We observe that mlp Context Bridge has more accurate edge predictions,  Mix\_FFN has more accurate segmentation results due to the supplement of local information, while our Enhanced Mix\_FFN gets better segmentation performance and moderate edge prediction because of the discriminative global and local features.

\subsection{Comparison with state-of-the-art methods}
This section reports the comparison results of MISSFormer and previous state-of-the-art methods on the Synapse dataset and ACDC dataset. 

\textbf{Experiment results on Synapse dataset.} Table 8 presents the comparison results of proposed MISSFormer and previous state-of-the-art methods. As shown in Table 8, the proposed method achieved state-of-the-art performance in almost all measures, and it is worth mentioning that the encoder of Transunet and Swin-Unet is pre-trained on ImageNet, while the MISSFormer trained on Synapse dataset from scratch, which indicates that MISSFormer capture the better long-range dependencies and local context to make strong feature representations. The visualization results are shown in Figure 6. It can be seen that our MISSFormer achieves better edge predictions and hard example segmentations compared to Tranunet and Swin-Unet, even in the bad case. Comparing MISSFormer and MISSFormer\_S, MISSFormer has precise results and less false segmentation because of the integration of multi-scale information. 

\setcounter{table}{8}
\begin{table}[t]
	\centering
	\begin{tabular}{l|c|ccc}
		\hline
		Methods & DSC$\uparrow$ & RV & Myo & LV \\
		\hline
		\hline
		%\cline{2-4}
		R50 U-Net&87.55&87.10& 80.63& 94.92	\\
		R50 Att-UNet& 86.75&87.58& 79.20& 93.47	\\
		R50 ViT& 87.57&86.07&81.88&94.75	\\
		TranUnet&89.71&88.86& 84.53&95.73\\
		SwinUnet& 90.00&88.55& 85.62& \textbf{95.83}	\\
		\hline
		MISSFormer& \textbf{90.86} & \textbf{89.55}  &\textbf{88.04}   &94.99\\
		\hline	
	\end{tabular}
	\caption{Comparison to state-of-the-art methods on ACDC dataset. The results of other experiments are original from Swin-Unet\cite{Cao2021}.}
	\label{table9}
\end{table}

\textbf{Experiment results on ACDC dataset.} We evaluate our method on the ACDC dataset in the form of MRI. Table 9 presents the segmentation accuracy. MISSFormer maintains the first position because of the powerful feature extraction, which indicates the outstanding generalization and robustness of MISSFormer. 

\section{Conclusion}
In this paper, we presented MISSFormer, a position-free and hierarchical U-shaped medical image segmentation transformer, which explored the global dependencies and local context capture. The proposed Enhanced Mix Block can overcome the problem of feature discrimination limitation caused by the direct embedding of convolution in feed-forward neural network effectively and make discriminative feature representations. Based on these core designs, we further investigated the integration of multi-scale features generated by our hierarchical transformer encoder, which is essential for accurate segmentation. We evaluated our method on two different forms of datasets, the superior results demonstrate the effectiveness and robustness of MISSFormer.

%\section{References.}
	\bibliography{000.bib}

\end{document}